\newif\iftechreport
\techreportfalse 

    \documentclass[10pt,journal,final,twoside,twocolumn]{IEEEtran}

\usepackage{amssymb}
\usepackage{amsmath}
\usepackage{mathtools}
\usepackage[caption=false,font=footnotesize]{subfig}
\usepackage{color}
\usepackage[none]{hyphenat}
\usepackage{booktabs}
\usepackage{multirow}
\usepackage{url}

\usepackage{breqn,ulem}
\usepackage{tikz}
	\usetikzlibrary{shapes,arrows}
\usepackage{pbox}
\usepackage{blindtext}
\usepackage{array}
\usepackage[ruled,vlined]{algorithm2e}
\DeclareMathOperator*{\argmin}{arg\,min}
\DeclareMathOperator{\prox}{\mathrm{prox}}
\usepackage{url}

\bstctlcite{IEEEexample:BSTcontrol}

    \title{Unmixing dynamic PET images with variable specific binding kinetics}

    \author{Yanna Cruz Cavalcanti,~\IEEEmembership{Student Member,~IEEE}, Thomas Oberlin,~\IEEEmembership{Member,~IEEE}, \\Nicolas Dobigeon,~\IEEEmembership{Senior Member,~IEEE}, Simon Stute, Maria Ribeiro, Clovis Tauber,~\IEEEmembership{Member,~IEEE}
    \thanks{Y. C. Cavalcanti, T. Oberlin and N. Dobigeon are with University of Toulouse, IRIT/INP-ENSEEIHT, CNRS, 2 rue Charles Camichel, BP 7122, 31071 Toulouse Cedex 7, France (e-mail: \{Yanna.Cavalcanti, Thomas.Oberlin, Nicolas.Dobigeon\}@enseeiht.fr).}
    \thanks{S. Stute is with UMRS Inserm U1023 IMIV-CEA SHFJ, 91400 Orsay, France (e-mail:
    simon.stute@cea.fr).}
    \thanks{M. Ribeiro and C. Tauber are with UMRS Inserm U930 - Universit\'e de Tours, 37032 Tours, France (e-mail:
    \{maria.ribeiro, clovis.tauber\}@univ-tours.fr).}
    \thanks{Part of this work has been supported by CAPES.}
    }

\begin{document}

    \maketitle

\bigskip
%
\begin{abstract}
\textit{Objective:} To analyze dynamic positron emission tomography (PET) images, various generic multivariate data analysis techniques have been considered in the literature, such as principal component analysis (PCA), independent component analysis (ICA), factor analysis and nonnegative matrix factorization (NMF). Nevertheless, these conventional approaches neglect any possible nonlinear variations in the time activity curves describing the kinetic behavior of tissues with specific binding, which limits their ability to recover a reliable, understandable and interpretable description of the data. This paper proposes an alternative analysis paradigm that accounts for spatial fluctuations in the exchange rate of the tracer between a free compartment and a specifically bound ligand compartment. \textit{Methods:} The method relies on the concept of linear unmixing, usually applied on the hyperspectral domain, which combines NMF with a sum-to-one constraint that ensures an exhaustive description of the mixtures. The spatial variability of the signature corresponding to the specific binding tissue is explicitly modeled through a perturbed component. \textit{Results:} The performance of the method is assessed on both synthetic and real data and is shown to compete favorably when compared to other conventional analysis methods. \textit{Conclusion:} The proposed method improved both factor estimation and proportions extraction for specific binding. \textit{Significance:} Modeling the variability of the specific binding factor has a strong potential impact for dynamic PET image analysis.
\end{abstract}

    \begin{IEEEkeywords}
    Dynamic PET image, unmixing, brain imaging, factor analysis, matrix factorization, NMF.
    \end{IEEEkeywords}
    \vspace{-0.5cm}

\section{Introduction}
\label{sec:introduction}
\IEEEPARstart{D}{ynamic} positron emission tomography (PET) is a non-invasive nuclear imaging technique that allows biological processes to be quantified and organ metabolic functions to be evaluated through the three-dimensional measure of the radiotracer concentration over time. 

The analysis of dynamic PET images, in particular the quantification of the kinetic properties of the tracer, requires the extraction of tissue time-activity-curves (TACs) in order to estimate the parameters from compartmental modeling \cite{Innis2007}. Nevertheless, PET images are corrupted by a prominent statistical noise and elementary TACs are mixed up due to the partial volume effect. Therefore, inferring rigorous and reliable information from these images still remains a challenging issue.

Several generic methods have been applied to estimate elementary TACs and their corresponding proportions from dynamic PET images. These techniques have different denominations depending on the application context, but all aim at tackling blind source separation (BSS) problems. For instance, Barber \cite{Barber1980} and Cavailloles et al. \cite{Cavailloles1984} proposed matrix factorization-based PET analysis techniques, referred to as factor analysis of dynamic structures (FADS). An improvement of FADS taking into account the nonnegative physiological characteristic of PET images 
was proposed subsequently by Wu et al. \cite{Wu1995} and Sitek et. al. \cite{Sitek2000}. Nonnegative matrix factorization (NMF) techniques pursue the same objective, under nonnegativity constraints on the latent factors to be recovered, and have been intensively applied in dynamic PET studies 
\cite{Lee2001a,Padilla2012,Schulz2013}.
The works of Sitek et al. \cite{Sitek2002} and El Fakhri et al. \cite{Elfakhri2005} improved nonnegative FADS with a penalization that promoted non-overlapping regions in each voxel.

The approach proposed in this paper follows the same line as NMF or nonnegative FADS. It aims at decomposing each PET voxel TAC into a weighted combination of pure physiological factors, representing the elementary TAC associated with the different tissues present within the voxel. This factor modeling is enriched with a sum-to-one constraint to the factor proportions, so that they can be interpreted as tissue percentages within each voxel. In particular, this additional constraint explicitly solves the scaling ambiguity inherent to any NMF models, which has proven to increase robustness as well as interpretability. This BSS technique, referred to as \emph{unmixing} or \emph{spectral mixture analysis}, originates from the geoscience and remote sensing literature \cite{Bioucas-Dias2012} and  has proven its interest in other applicative contexts, such as microscopy \cite{Huang2011plos} and genetics \cite{Dobigeon2012ultra}.

The linearity assumption for voxel decomposition underlined by the above-mentioned PET analysis methods can be envisioned in the light of compartment modeling, a tool widely employed to describe the kinetic behavior occurring within the voxel. The selection of a given model should be based on the radiotracer under study \cite{Gunn1997,Innis2007} but most of the models consider that the measured signal in a given voxel is the sum of the comprising compartments.

\begin{figure}
\centering
\begin{tikzpicture}
\tikzstyle{cfns}=[minimum width=2cm,minimum height=2cm,rectangle,draw=blue,rounded corners=4pt,fill=blue!40,text=white,text width = 2cm,align=center]
\tikzstyle{cp}=[minimum width=1cm,minimum height=5cm,cylinder,shape border rotate=90,draw=red,fill=red!40,text=white,text width = 1cm,align=center]
\tikzstyle{cs}=[minimum width=2cm,minimum height=2cm,rectangle,draw=blue,rounded corners=4pt,fill=blue!70,text=white,text width = 2cm,align=center]
\tikzstyle{tissue}=[minimum width=7.3cm,minimum height=3cm,rectangle,draw=black,dashed,rounded corners=4pt]
\tikzstyle{suite}=[->,>=stealth,thick,rounded corners=4pt]
\node[cfns] (cfns) at (0,0) {\Large{$C_{F+NS}$}};
\node[cs] (cs) at (3,0) {\Large{$C_S$}};
\node[cp] (cp) at (-3,0) {\Large{$C_P$}};
\node[tissue,label=above:{Tissue}] (tissue) at (1,0) {};
\draw[->,dashed] (-2.4,0.5) -- (-1.1,0.5) node[midway,sloped,above] {$k_1$};
\draw[->,dashed] (-2.4,-0.5) -- (-1.1,-0.5) node[midway,sloped,above] {$k_2$};
\draw[->,dashed] (1.2,0.5) -- (1.9,0.5) node[midway,sloped,above] {$k_3$};
\draw[->,dashed] (1.2,-0.5) -- (1.9,-0.5) node[midway,sloped,above] {$k_4$};
\end{tikzpicture}
\caption{Configuration of the classic three-compartment kinetic model used in many imaging studies.}
\label{fig:2tiss}
\end{figure}
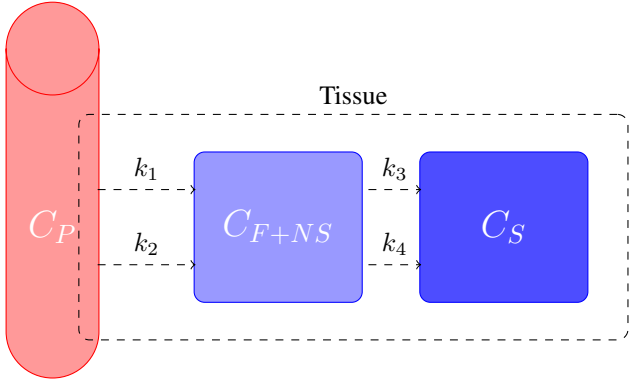

However, factor TACs to be recovered cannot always be assumed to have constant kinetic patterns, as implicitly considered in conventional methods. Figure \ref{fig:2tiss} depicts an example of a 2-tissue compartmental model, where the radioligand is assumed to move between three compartments:  $C_p$ represents the radioligand concentration in arterial plasma, $C_{F+NS}$ represents the free plus non-specific compartment, and $C_S$ represents the specifically bound compartment. The exchange between compartments are subject to rate constants $k_j$ ($j=1,\ldots,4$) \cite{Häggström2016}. Considering both the 2-tissue and reference compartment models, the assumption of constant kinetic patterns seems appropriate for the blood compartment as well as non-specific binding tissues, since they present some homogeneity besides some perfusion difference (e.g. white matter versus gray matter). Therefore their contribution to the voxel TAC should be fairly proportional to the fraction of this type of tissue in the voxel. However, things get different regarding the specific binding class, as the TAC associated with this tissue is nonlinearly dependent on both the perfusion and the concentration of the radiotracer target. The spatial variation on target concentration is in part governed by differences in the $k_3$ and $k_4$ kinetic parameters, which nonlinearly modify the shape of the TAC characterizing this particular class.  Muzi et al. \cite{Muzi2005} discussed the accuracy of parameter estimates for tumor regions and underlined high errors for the parameters related to specific binding, namely 26\% for $k_3$ and 49\% for $k_4$. These results were further confirmed by Schiepers et al. \cite{Schiepers2007}. More specifically, they studied the kinetics of lesioned regions that were tumor and treatment change predominant, showing that variations on $k_3$ and $k_4$ may allow for differentiation. Bai et al. \cite{Bai2013}  further discussed  nonuniformity in intratumoral uptake and its impact on predicting treatment response and tumor aggressiveness. Nonetheless, this fluctuation phenomenon has not been taken into account by the decomposition models from the literature.

The main motivation of this paper is to propose a more accurate description of the tissues and kinetics composing the voxels in dynamic PET images, in particular for those affected by specific binding. To this end, this work proposes to explicitly model the nonlinear variability inherent to the TAC corresponding to specific binding, by allowing the corresponding factor to vary spatially. This variation is approximated by a linear expansion over the atoms of a dictionary, which have been learned beforehand by conducting a principal component analysis on a learning dataset. Note that part of this work has been previously presented at the European Signal Processing Conference (EUSIPCO) \cite{Cavalcanti2017eusipco}.

The sequel of this paper is organized as follows. The proposed mixing-based analysis model is described in Section \ref{sec:prob_stat}. Section \ref{sec:palm} presents the corresponding unmixing algorithm able to recover the factors, their corresponding proportions in each voxel and the variability maps. Simulation results obtained with synthetic data are reported in Section \ref{sec:synt_sim}. Experimental results on real data are provided in Section \ref{sec:real_sim}. Section \ref{sec:concl} concludes the paper.

\section{Problem Statement}
\label{sec:prob_stat}
\subsection{Specific binding linear mixing model (SLMM)}
\label{subsec:plmm}

Consider $N$ voxels of a 3D dynamic PET image acquired at $L$ successive time-frames. First, we omit the spatial blurring induced by the point spread function (PSF) of the instrument and any measurement noise. The TAC in the $n^{th}$ voxel  ($n\in\left\{1,\ldots,N	\right\}$) over the $L$ time-frames is denoted $\mathbf{x}_n=[x_{1,n},\ldots,x_{L,n}]^T$. Akin to various BSS techniques and following the linear mixing model (LMM) for instance advocated in the PET literature by Barber et al. \cite{Barber1980}, each TAC $\mathbf{x}_n$ is assumed to be a linear combination of $K$ \emph{elementary factors} $\mathbf{m}_k$\vspace{-0.25cm}
\begin{equation}
\mathbf{x}_n = \sum_{k=1}^{K}{\mathbf{m}_k a_{k,n}}\vspace{-0.25cm}
\label{eq:lmm}
\end{equation}
where  $\mathbf{m}_k=[m_{1,k},\ldots,m_{L,k}]^T$ denotes the {pure TAC of the $k^{th}$ tissue type and $a_{k,n}$ is the factor proportion of the $k^{th}$ tissue in the $n^{th}$ voxel. The factors $\mathbf{m}_k$ ($k=1,\ldots,K$) correspond to the kinetics of the radiotracer in a particular type of tissue in which they are supposed spatially homogeneous. For instance, the experiments conducted in this work and described in Sections \ref{sec:synt_sim} and \ref{sec:real_sim} consider $3$ types of tissues that fall into this category: the blood, the non-specific gray matter and the white matter.

Additional constraints regarding these sets of variables are assumed. First, since the elementary TACs are expected to be nonnegative, the factors are constrained as\vspace{-0.15cm}
\begin{equation}
{m}_{l,k} \geq 0,\quad \forall l,k. \vspace{-0.15cm}
\label{eq:constrM}
\end{equation}
Moreover, nonnegativity and sum-to-one constraints are assumed for all the factor proportions ($n=1,\ldots,N$)
\begin{equation}
\forall k\in\left\{1,\ldots,K\right\},\ {a}_{k,n} \geq 0 \quad \text{and} \quad
\sum_{k=1}^K {a}_{k,n}  = 1.
\label{eq:constrA}
\end{equation}
For a given voxel indexed by $n$, this sum-to-one constraint \eqref{eq:constrA} enforces the mixing coefficients $a_{k,n}$ ($k=1,\ldots,K$) to be interpreted as concentrations \cite{Keshava2003}. Therefore, all factor proportion vectors $\mathbf{a}_n$ ($n=1,\ldots,N$) lie inside the unit $(K-1)$-simplex. Similarly, the TAC $\mathbf{x}_n$ belongs to the convex set whose vertices are the columns of $\mathbf{M}$, represented by a $(K-1)$-simplex in $\mathbb{R}^L$ \cite{Bioucas-Dias2012}. Fig. \ref{fig:simplex} shows an example of a simplex defined by $K=3$ elementary factors.

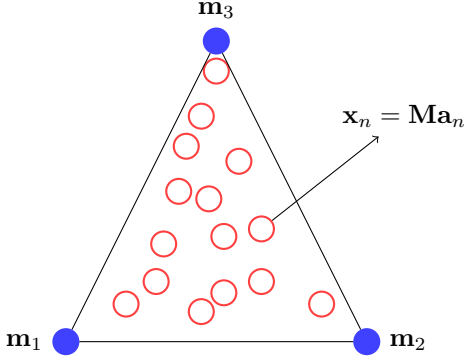
\begin{figure}
\centering
\begin{tikzpicture}
\tikzstyle{u} = [circle, thick, draw=blue!75,fill=blue!75]
\tikzstyle{n} = [circle, thick, draw=red!75]
\tikzstyle{suite}=[->]
\node (y) at (2.5,1) {$\mathbf{x}_n=\mathbf{Ma}_n$};
\node[u, label=left:$\mathbf{m}_1$] (m1) at (-2,-2) {};
\node[u, label=right:$\mathbf{m}_2$] (m2) at (2,-2) {};
\node[u, label=above:$\mathbf{m}_3$] (m3) at (0,2) {};
\node[n] (y1) at (0,1.6) {};
\node[n] (y2) at (-0.2,1) {};
\node[n] (y3) at (-0.4,0.6) {};
\node[n] (y4) at (0.3,0.4) {};
\node[n] (y5) at (-0.5,0) {};
\node[n] (y6) at (-0.1,-0.1) {};
\node[n] (y7) at (0.6,-0.5) {};
\node[n] (y8) at (-0.7,-0.7) {};
\node[n] (y9) at (0.1,-0.6) {};
\node[n] (y10) at (-0.8,-1.2) {};
\node[n] (y11) at (-1.2,-1.5) {};
\node[n] (y12) at (0.6,-1.2) {};
\node[n] (y13) at (0.1,-1.35) {};
\node[n] (y14) at (-0.2,-1.6) {};
\node[n] (y15) at (1.4,-1.5) {};
%
\textbf{\draw (m1) -- (m2);}
\textbf{\draw (m2) -- (m3);}
\textbf{\draw (m3) -- (m1);}
\textbf{\draw[suite] (y7) -- (y);}
\end{tikzpicture}
\caption{Illustration of the simplex for a mixing matrix of 3 factors. The blue circles represent the vertices of the simplex, corresponding to the factors and red circles are the TACs.}
\label{fig:simplex}
\end{figure}

More importantly, when factors are affected by possibly nonlinear and spatially varying fluctuations within the image, the conventional NMF-like linear mixing model \eqref{eq:lmm} no longer provides a sufficient description of data. Therefore, over recent years, factor variability has received increased interest in the hyperspectral imagery literature as it allows changes on lightening and environment to be taken into account \cite{Zare2014,Halimi2015}. Recently, Thouvenin et al. have proposed a perturbed LMM (PLMM) to further address this problem \cite{Thouvenin2016}. In the dynamic PET image framework, factor variability is expected to mainly affect the TAC associated with specific binding, denoted $\mathbf{m}_1$, while the possible variabilities in the TACs $\mathbf{m}_k$ ($k\in\left\{2,\ldots,K\right\}$) related to tissues devoid of a specifically bound compartment are supposed weaker and neglected in this study. Since this so-called specific binding factor (SBF) is assumed to vary spatially, it will be spatially indexed.
Thus, adapting the PLMM approach to our problem, the SBF} in a given voxel will be modeled as a spatially-variant additive perturbation affecting a nominal and common SBF $\bar{\mathbf{m}}_1$\vspace{-0.15cm}
\begin{equation}
\label{eq:plmm}
\mathbf{m}_{1,n} = \bar{\mathbf{m}}_1 + \delta\mathbf{m}_{1,n}\vspace{-0.15cm}
\end{equation}
where the additive term $\delta\mathbf{m}_{1,n}$ describes its spatial variability over the image.  However, recovering the spatial fluctuation $\delta\mathbf{m}_{1,n}$ in each image voxel is a high-dimensional problem. To reduce this dimensionality, the variations will be assumed to lie inside a subspace of small dimension $N_v \ll L$.
As a consequence, similarly to the strategy followed in \cite{Park2014}, the additive terms $\delta\mathbf{m}_{1,n}$ ($n\in\left\{1,\ldots,N\right\}$) are supposed to be approximated by the linear expansion\vspace{-0.15cm}
\begin{equation}
\label{eq:m1n}
 \delta\mathbf{m}_{1,n} = \sum_{i=1}^{N_v} b_{i,n} {\mathbf{v}_i},\vspace{-0.15cm}
\end{equation}
 where the  $N_v$ variability basis elements $\mathbf{v}_1,\ldots,\mathbf{v}_{N_v}$  can be chosen beforehand, e.g., by conducting a PCA on a learning set composed of simulated or measured SBFs. The PCA aims at extracting the main variability patterns, while allowing for dimension reduction. Thus, the set of coefficients $\left\{b_{1,n},\ldots,b_{N_v,n}\right\}$ quantify the amount of variability in the $n^{th}$ voxel. The nominal SBF $\bar{\mathbf{m}}_1$ is also roughly estimated from the lowest values of this dataset and further fixed so that all the variability coefficients are nonnegative, in order to reduce correlation between the variability elements and the other tissue factors.

Combining the linear mixing model \eqref{eq:lmm}, the perturbation model \eqref{eq:plmm} and its linear expansion \eqref{eq:m1n}, the voxel TACs are described according to the following so-called   specific binding linear mixing model ({SLMM)\vspace{-0.15cm}
\begin{equation}\label{eq:model_with_variability}
\mathbf{x}_n = a_{1,n} {\bigg(\mathbf{\bar{m}}_1+\sum_{i=1}^{N_v}{b_{i,n} \mathbf{v}_i }\bigg) }+\sum_{k=2}^{K}{a_{k,n} \mathbf{m}_k }.\vspace{-0.15cm}
\end{equation}
To be fully comprehensive and motivated by the findings in \cite{Henrot2014}, this work also proposes to explicitly model the PET scan point spread function (PSF), combining a deconvolution step jointly with parameter estimation. We will denote by $\mathbf{H}$ the linear operator that computes the 3D convolution by some known and  spatially invariant PSF, which leads to\vspace{-0.15cm}
\begin{equation}
\mathbf{Y} = \mathbf{MAH} +     \underbrace{\Big[\mathbf{E}_1\mathbf{A}\circ \mathbf{V}\mathbf{B}\Big]\mathbf{H}}_{\Delta} +\mathbf{R}\vspace{-0.15cm}
\label{eq:plmm_matrix}
\end{equation}
where  $\mathbf{M} = [\mathbf{\bar{m}}_1,...,\mathbf{m}_{k}]$ is a $L \times K$ matrix containing the factor TACs, $\mathbf{A} = \left[\mathbf{a}_1,\ldots,\mathbf{a}_n\right]$ is a $K \times N$ matrix composed of the factor proportion vectors, ``$\circ$'' is the Hadamard point-wise product, $\mathbf{E}_{1}$ is the matrix $[\mathbf{1}_{L,1} \mathbf{0}_{L,K-1}]$, $\mathbf{V}= [\mathbf{v}_1,\ldots,\mathbf{v}_{N_v}]$ is the $L \times N_v$ matrix containing the basis elements used to expand the spatial variability of the SBF, $\mathbf{B} = \left[\mathbf{b}_1,\ldots,\mathbf{b}_n\right]$ is the $N_v \times N$ matrix containing the intrinsic proportions, and $\mathbf{R}=\left[\mathbf{r}_1,\ldots,\mathbf{r}_N\right]^T$ is the $L \times N$ matrix accounting for noise and mismodeling. Note that if $\mathbf{B} = \mathbf{0}$ and $\mathbf{H} = \mathbf{I}$, the model in  (\ref{eq:plmm_matrix}) reduces to the conventional linear mixing model generally assumed by factor model techniques like NMF and ICA.


While noises associated with count rates are traditionally modeled by a Poisson distribution \cite{Shepp1982}, postprocessing corrections and filtering operated by modern PET systems significantly alter the nature of the noise corrupting the final reconstructed images. Modeling the noise on this final data is a highly challenging task \cite{Wilson1994}. However, as demonstrated in \cite{Fessler1994}, pre-corrected PET data can be sufficiently approximated by a Gaussian distribution. As a consequence, in this work, the noise vectors $\mathbf{r}_n = [r_{1,n},\ldots,r_{L,n}]$ ($n\in\left\{1,\ldots,N\right\}$) are assumed to be normally distributed. Moreover, without loss of generality, all vector components $r_{\ell,n}$ ($\ell=1,\ldots,L$ and $n=1,\ldots,N$) will be assumed to be independent and identically distributed. This assumption seems to evade any spatial and temporal correlations that may characterize the noise generally affecting the reconstructed PET images \cite{Tichy2015}. However, the proposed model can be easily generalized to handle colored noise by weighting the model discrepancy measure according to the noise covariance matrix, as in \cite{Fessler1994}. Alternatively, after diagonalizing the noise covariance matrix, the PET image to be analyzed can undergo a conventional whitening pre-processing step \cite{Thireou2006,Bullmore2001,Turkheimer2003}.

In addition to the nonnegativity constraints applied to the elementary factors \eqref{eq:constrM} and factor proportions \eqref{eq:constrA}, the intrinsic variability proportion matrix $\mathbf{B}$ is also assumed to be nonnegative, mainly to avoid spurious ambiguity, i.e.,
\vspace{-0.05cm}
\begin{equation}
\mathbf{B} \succeq \boldsymbol{0}_{N_v,N},\vspace{-0.15cm}
\label{eq:var_constr}
\end{equation}
where $\boldsymbol{0}_{N_v,N}$ denotes the $N_v\times N$-matrix made of $0$'s and $\succeq$ stands for a component-wise inequality. We accordingly fix the nominal SBF $\mathbf{\bar{m}}_1$ with a robust estimation of the TAC chosen as a lower bounding signature of a set of previously generated or measured SBF TACs. This means that a negative bias on the SBF is artificially introduced to model the spatially-varying SBF TACs $\mathbf{{m}}_{1,n}$ ($n\in \left\{1,\ldots,N\right\}$). This is alternatively compensated by a variability that is distorted by the same quantity but positively. This constraint is chosen to avoid a high correlation between the other factor TACs and $\sum_{i=1}^{N_v}{\mathbf{v}_ib_{i,n}}$ when $b_{i,n}$ is allowed to be negative. Capitalizing on this model, the unmixing-based analysis of dynamic PET images is formulated in the next paragraph.

\subsection{Problem formulation}
\label{subsec:probform}

The SLMM (\ref{eq:plmm_matrix}) and constraints (\ref{eq:constrM}), (\ref{eq:constrA}) and (\ref{eq:var_constr}) can be combined to formulate a constrained optimization problem. In order to estimate the matrices $\mathbf{M}$, $\mathbf{A}$, $\mathbf{B}$, a proper cost function is defined. The data-fitting term is defined as the  Frobenius norm $\| \cdot\|_F^2$ of the difference between the dynamic PET image $\mathbf{Y}$ and the proposed data modeling $\mathbf{MAH}+\Delta$. This corresponds to the negative log-likelihood under the assumption of Gaussian noise. Since the problem is ill-posed and non-convex, additional regularizers become essential. In this paper, we propose to define penalization functions $\Phi$, $\Psi$ and $\Omega$ to reflect the available \textit{a priori} knowledge on $\mathbf{M}$, $\mathbf{A}$ and $\mathbf{B}$, respectively. The optimization problem is then defined as\vspace{-0.10cm}
\begin{equation}
(\mathbf{M}^*, \mathbf{A}^*, \mathbf{B}^*) \in \argmin_{\mathbf{M}, \mathbf{A}, \mathbf{B}} \Big\{\mathcal{J}(\mathbf{M}, \mathbf{A}, \mathbf{B}) \text{ s.t. (\ref{eq:constrM}),(\ref{eq:constrA}),(\ref{eq:var_constr})}\Big\}\vspace{-0.10cm}
\label{eq:optprob}
\end{equation}
with\vspace{-0.10cm}
\begin{equation}
\begin{split}
\mathcal{J}(\mathbf{M}, \mathbf{A}, \mathbf{B})=\frac{1}{2} \left\| \mathbf{Y}-\mathbf{MAH}-\Big[\mathbf{E}_1\mathbf{A}\circ \mathbf{V}\mathbf{B})\Big]\mathbf{H}\right\|_F^2\\
+\alpha\Phi(\mathbf{A})+\beta\Psi(\mathbf{M})+\lambda\Omega(\mathbf{B}) \vspace{-0.10cm}
\label{eq:costfct}
\end{split}
\end{equation}
where the parameters $\alpha$, $\beta$ and $\lambda$ control the trade-off between the data fitting term and the penalties $\Phi(\mathbf{A})$, $\Psi(\mathbf{M})$ and $\Omega(\mathbf{B})$, described hereafter.

\subsubsection{Factor proportion penalization}
\label{subsubsec:abundpen}
The factor proportions representing the amount of different tissues are assumed to be spatially smooth, since neighboring voxels may contain the same tissues. We thus penalize the energy of the spatial gradient
\begin{equation}
\Phi(\mathbf{A}) = \frac{1}{2}\|\mathbf{AS}\|_F^2,
\label{eq:abundpen2}
\end{equation}
where $\mathbf{S}$ is the operator computing the first-order spatial finite differences.  More details are reported in \cite{Cavalcanti2017TR}.

\subsubsection{Factor penalization}
\label{subsubsec:endmpen}
The chosen factor penalization benefits from the availability of rough factor  TACs  estimates $\mathbf{M}^0 = \left[\bar{\mathbf{m}}_1^0,\ldots,\mathbf{m}_K^0\right]$. Thus, we propose to enforce similarity (in term of mutual Euclidean distances) between these primary estimates and the factor TACs to be recovered\vspace{-0.15cm}
\begin{equation}
\begin{aligned}
\Psi(\mathbf{M})&=\frac{1}{2}{\big\|\mathbf{M}-\mathbf{M}^0\big\|_F^2}.
\label{eq:endmpen2}
\end{aligned}
\end{equation}

\subsubsection{Variability penalization}
\label{subsubsec:variabilitypen}
The SBF variability is expected to affect only a small number of voxels, those belonging to the region containing the SBF. As a consequence,  we propose to enforce sparsity via the use of the $\ell_1$-norm, also known as the LASSO regularizer \cite{Tibshrani1996}\vspace{-0.10cm}
\begin{equation}
\Omega(\mathbf{B}) = \|\mathbf{B}\|_1\vspace{-0.10cm}
\label{eq:varpen2}
\end{equation}
where $\|.\|_1$ is the $\ell_1$ norm. This penalty forces $ b_{i,n}$ to be $0$ outside the high-uptake region, thus reducing overfitting.

\section{A PALM-based algorithm}
\label{sec:palm}
Given the nature of the optimization problem (\ref{eq:optprob}), which is genuinely nonconvex and nonsmooth, the adopted minimization strategy relies on the proximal alternating linearized minimization (PALM) scheme \cite{Bolte2013}. PALM is an iterative, gradient-based algorithm which generalizes the Gauss-Seidel method. It consists in iterative proximal gradient steps with respect to $\mathbf{A}$, $\mathbf{M}$ and $\mathbf{B}$ and ensures convergence to a local critical point $\mathbf{A}^*$, $\mathbf{M}^*$ and $\mathbf{B}^*$.
    The principle of PALM is briefly recalled in \cite{Cavalcanti2017TR}. It is specifically instantiated for the unmixing-based kinetic component analysis considered in this paper. The resulting SLMM unmixing algorithm, whose main steps are described in the following paragraphs, is summarized in Algo. \ref{algo:globalplmm}. 

    \LinesNumbered
    \begin{algorithm}
    \DontPrintSemicolon
    \KwData{$\mathbf{Y}$}
    \KwIn{$\mathbf{A}^{0}$, $\mathbf{M}^{0}$, $\mathbf{B}^{0}$}
    $k \leftarrow 0$\;
    \While{stopping criterion not satisfied}{
    \label{algostep:M} $\mathbf{M}^{k+1} \leftarrow \mathcal{P}_{+}\bigg( \mathbf{M}^{k}-\frac{\gamma}{L_M^{k}} \nabla_{\mathbf{M}}\mathcal{J}(\mathbf{M}^{k}, \mathbf{A}^{k+1}, \mathbf{B}^{k})\bigg)$ \;
    \label{algostep:A} $\mathbf{A}^{k+1} \leftarrow \mathcal{P}_{\mathcal{A}_R}\bigg( \mathbf{A}^{k}-\frac{\gamma}{L_A^{k}} \nabla_{\mathbf{A}}\mathcal{J}(\mathbf{M}^{k}, \mathbf{A}^{k}, \mathbf{B}^{k})\bigg)$ \;
    \label{algostep:B} $\mathbf{B}^{k+1} \leftarrow \newline
    \prox_{\frac{\lambda}{L_B^{k}}\|.\|_1}\bigg(\mathcal{P}_{+}\bigg(\mathbf{B}^{k}-\nolinebreak\frac{\gamma}{L_B^{k}} \nabla_{\mathbf{B}}\mathcal{J}(\mathbf{M}^{k+1}, \mathbf{A}^{k+1}, \mathbf{B}^{k})\bigg)\bigg)$ \;
    $k \leftarrow k+1$\;
    }
    $\mathbf{A} \leftarrow \mathbf{A}^{k+1}$\;
    $\mathbf{M} \leftarrow \mathbf{M}^{k+1}$\;
    $\mathbf{B} \leftarrow \mathbf{B}^{k+1}$\;
    \KwResult{$\mathbf{A}$, $\mathbf{M}$, $\mathbf{B}$}
    \caption{SLMM unmixing: main algorithm \label{algo:globalplmm}}
    \end{algorithm}

\subsection{Optimization with respect to $\mathbf{M}$}
\label{subsec:optM}

A direct application of \cite{Bolte2013} under the constraints defined by (\ref{eq:constrM}) leads to the following updating rule
\begin{equation}
\mathbf{M}^{k+1} = \mathcal{P}_{+}\bigg( \mathbf{M}^{k}-\frac{1}{L_M^{k}} \nabla_{\mathbf{M}}\mathcal{J}(\mathbf{M}^{k}, \mathbf{A}^{k+1}, \mathbf{B}^{k})\bigg)
\label{Mpalm}
\end{equation}
where $\mathcal{P}_{+}(\cdot)$ is the projector onto the nonnegative set $\{\mathbf{X}|\mathbf{X}\succeq \mathbf{0}_{L,R}\}$ and the required gradient writes
\begin{equation}
\begin{split}
\nabla_{\mathbf{M}}\mathcal{J}(\mathbf{M}, \mathbf{A}, \mathbf{B}) = \left(\left(\mathbf{E}_1\mathbf{A}\circ \mathbf{V}\mathbf{B}\right)\mathbf{H}-\mathbf{Y}\right)\mathbf{H}^T\mathbf{A}^T \\
+\mathbf{M}(\mathbf{AH}\mathbf{H}^T\mathbf{A}^T)+\beta(\mathbf{M}-\mathbf{M}^0).
\label{Mgradient}
\end{split}
\end{equation}

\subsection{Optimization with respect to $\mathbf{A}$}
\label{subsec:optA}
Similarly to paragraph \ref{subsec:optM}, the factor proportion update is defined as the following
\begin{equation}
\mathbf{A}^{k+1} = \mathcal{P}_{\mathcal{A}_R}\bigg( \mathbf{A}^{k}-\frac{1}{L_A^{k}} \nabla_{\mathbf{A}}\mathcal{J}(\mathbf{M}^{k}, \mathbf{A}^{k}, \mathbf{B}^{k})\bigg),
\label{Apalm}
\end{equation}
where $\mathcal{P}_{\mathcal{A}_R}(\cdot)$ is the projection on the set $\mathcal{A}_R$ defined by the factor proportion constraints (\ref{eq:constrA}), which can be computed with efficient algorithms, see, e.g., \cite{Condat2015}. The gradient can be computed as
\begin{equation*}
\nabla_{\mathbf{A}}\mathcal{J}(\mathbf{M}, \mathbf{A}, \mathbf{B}) = -\mathbf{M}^T(\mathbf{D}_A)-\mathbf{E}_1^T(\mathbf{D}_A \circ \mathbf{V}\mathbf{B})+\alpha\mathbf{A}\mathbf{S}\mathbf{S}^T
\label{Agradient}
\end{equation*}
with $\mathbf{D}_A = (\mathbf{Y}-\mathbf{M}\mathbf{A}\mathbf{H}-(\mathbf{E}_1\mathbf{A}\circ \mathbf{V}\mathbf{B})\mathbf{H})\mathbf{H}^T$.

\subsection{Optimization with respect to $\mathbf{B}$}
\label{subsec:optbeta}

Finally, the updating rule for the variability coefficients can be written as
\begin{equation*}
\mathbf{B}^{k+1} =  \prox_{\frac{\lambda}{L_B^{k}}\|.\|_1}\bigg(\mathcal{P}_{+}\bigg(\mathbf{B}^{k}-\frac{1}{L_B^{k}} \nabla_{\mathbf{B}}\mathcal{J}(\mathbf{M}^{k+1}, \mathbf{A}^{k+1}, \mathbf{B}^{k})\bigg)\bigg),
\label{betapalm}
\end{equation*}

\section{Evaluation on Synthetic Data}
\label{sec:synt_sim}
\subsection{Synthetic data generation}
To illustrate the accuracy of our algorithm, experiments are first conducted on synthetic data for which the ground truth of the main parameters of interest (i.e., factor TACs and factor proportion maps) is known. In the clinical PET framework, ground truth concerning the tracer kinetics and uptake is never completely known. Meanwhile, simulations benefit from an entire knowledge of the patient properties and kinetics, and their degree of complexity and details can be selected according to the purpose of the study. Furthermore, several simulations can be performed in a reasonable time.

    Thus, experimentations are conducted on set of $20$ synthetic images of size $128 \times 128\times 64$. In these images, each voxel is constructed as a combination of $K=4$ pure TACs representative of the brain, which is the organ of interest in the present work: specific gray matter, pure blood or veins, pure white matter and non-specific gray matter. First, a high resolution dynamic PET numerical phantom with labeled regions of interest (ROIs) \cite{Stute2015}, has been used to create the ground truth for factor and proportions. In this phantom, all the distributions of the tracer per region have been extracted from a real dynamic PET acquisition and are observed in $L=20$ times of acquisition ranging from  $1$ to $5$ minutes in a total period of $60$ minutes.  Using a phantom extracted from real-life acquisitions brings to the synthetic image the complexity of real data. For instance, the pure TACs constitutive of this phantom image are created from the averaging of visually segmented regions and therefore are still mixed up due to partial volume. To simulate realistic variability of the SBF, a set of synthetic TACs generated through a realistic compartment-based model is used. More details can be found in \cite{Cavalcanti2017TR}.

    The overall generation process is as follows:
\begin{itemize}
\item The dynamic PET phantom \iftechreport showed in Fig. \ref{fig:gt_phantom} \fi has been first linearly unmixed using the N-FINDR \cite{Winter1999spie} and SUnSAL \cite{Bioucas2010} algorithms to select the ground-truth non-specific factor TAC $\mathbf{m}_2,...,\mathbf{m}_K$ and factor proportions $\mathbf{a}_1,\ldots,\mathbf{a}_N$, respectively. These factor TACs and corresponding factor proportion maps are depicted in Fig. \ref{fig:A_noise15} (left) and Fig. \ref{fig:M_noise15} (left), respectively.
\item Following a 2-tissue compartment-based model \cite{Phelps1986}, a large database of SBF TACs has been generated by randomly varying the $k_3$ parameter (representing the specific binding rate of the radiotracer in the tissue). A PCA was conducted on this dataset, and an analysis of the eigenvalues led to the choice of a unique variability basis element $\mathbf{V} = \mathbf{v}_1$ (i.e., $N_v=1$), depicted in Fig. \ref{fig:var} (left).
\item The nominal SBF TAC $\bar{\mathbf{m}}_1$ is then chosen as the TAC of minimum area under the curve (AUC) among all the TACs of this database. This TAC is depicted in Fig. \ref{fig:var} (right, red curve).
\item The $1$st row of the factor proportion matrix $\mathbf{A}$, namely $\mathbf{A}_1 \triangleq \left[a_{1,1},\ldots,a_{1,N}\right]$ was designed to locate the region associated with specific binding. Then, the $N_v\times N$ matrix $\mathbf{B} = [b_1,\ldots,b_N]$ mapping the SBF variability in each voxel has been artificially generated. The high-uptake region was divided into $4$ subregions with non-zero coefficients $b_n$, as shown in Fig. \ref{fig:B_noise15} (left), while these coefficients are set to $b_n=0$ outside the region affected with SBF. In each of these subregions, the non-zero coefficients $b_n$ have been drawn according to Gaussian distributions with a particular mean value and small variances.  The spatially-varying SBFs in each region are then generated according to the model in \eqref{eq:m1n} and \eqref{eq:plmm}. Some resulting typical SBF TACs are showed in Fig. \ref{fig:var}.
\end{itemize}

\begin{figure}[h!]
\begin{center}
\includegraphics[width=0.45\columnwidth]{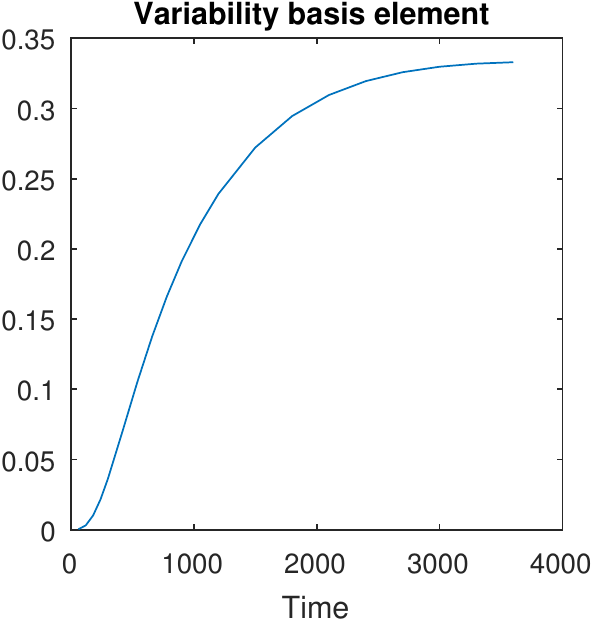}
\includegraphics[width=0.465\columnwidth]{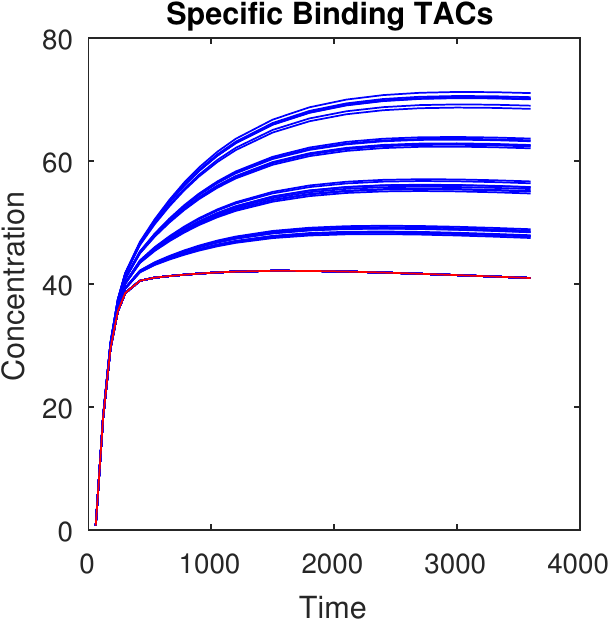}
\caption{Left: variability basis element $\mathbf{v}_1$ identified by PCA. Right: generated SBFs (blue) and the nominal SBF signature (red).}
\label{fig:var}
\end{center}
\end{figure}

After this primary generation process, a PSF defined as a space-invariant and isotropic Gaussian filter with FWHM$=4.4$mm is applied to the output image. In brain imaging using a clinical PET scanner, this is an acceptable approximation, since the degradation of the scanner resolution mainly affects the borders of the field-of-view \cite{Rahmin2013b,Mehranian2017}. Finally the measurements have been corrupted by a zero-mean white Gaussian noise with a signal-to-noise ratio $\mathrm{SNR}=15$dB, in agreement with a preliminary study conducted on the realistic replicas of \cite{Stute2015}, which showed that the SNR ranges from approximately $10$dB on the earlier frames to $20$dB on the latter ones.

\subsection{Compared methods}
The results of the proposed algorithm have been compared to those obtained with several classical linear unmixing methods and other BSS techniques. The methods are recalled below with their most relevant implementation details.

\paragraph{NMF (no variability)}  The NMF algorithm herein applied is based on multiplicative update rules using the Euclidean distance as cost function \cite{LeeNMF}. The stopping criterion is set to $10^{-3}$. To obtain a fair comparison mitigating scale ambiguity inherent to matrix factorization-like problem, results provided by the NMF have been normalized by the maximum value for the abundance, i.e.,
\begin{equation}
{\hat{\mathbf{A}}_{k}} \leftarrow \frac{{\hat{\mathbf{A}}_{k}}}{\big\|{\hat{\mathbf{A}}_{k}}\big\|_{\infty}} \quad \hat{\mathbf{m}}_k \leftarrow \hat{\mathbf{m}}_k \big\|{{\hat{\mathbf{A}}_{k}}}\big\|_{\infty}
\end{equation}
where ${\hat{\mathbf{A}}_{k}}$ denotes the $k$th row of the estimated factor proportion matrix ${\hat{\mathbf{A}}}$.

\paragraph{VCA (no variability)} The factor TACs are first extracted using the vertex component analysis (VCA) which requires pure voxels to be present in the analyzed images \cite{Nascimento2005}. The factor proportions are subsequently estimated by sparse unmixing by variable splitting and augmented Lagrangian (SUnSAL) \cite{Bioucas2010}.
\paragraph{LMM (no variability)} To appreciate the interest of explicitly modeling the spatial variability of the SBF, a depreciated version of the proposed SLMM algorithm is considered. More precisely, it uses the LMM \eqref{eq:lmm} without allowing the SBF $\mathbf{m}_{1,n}$ to be spatially varying. The stopping criterion, defined as $\varepsilon$, is set to $10^{-3}$. The values of the regularization parameter are reported in Table \ref{table:par_noise15}.
\paragraph{SLMM (proposed approach)} As detailed in Section \ref{subsec:plmm}, matrix $\mathbf{B}$ is constrained to be nonnegative to increase accuracy. Consequently, the nominal SBF TAC $\bar{\mathbf{m}}_1$ is initialized as the TAC with the minimum AUC learned from the generated database to ensure a positive $\mathbf{B}$. The regularization parameters have been tuned to the values reported in Table \ref{table:par_noise15}. As for the other approaches, the stopping criterion is set to $10^{-3}$.

Since the addressed problem is non-convex, these algorithms require an appropriate initialization. In this work, the factor TACs have been initialized as the outputs $\mathbf{M}^0$ of a K-means clustering conducted on the PET image. These K-means TACs estimates are also considered for performance comparison.

\begin{table}[h!]
\caption{factor proportion, factor and variability penalization parameters for LMM and SLMM with SNR$=15$dB}
\label{table:par_noise15}
\centering
\begin{tabular}{|c|c|c|}
\hline
 & LMM & SLMM\\
\hline
$\alpha$ & 0.010 & 0.010 \\
\hline
$\beta$ & 0.010 & 0.010 \\
\hline
$\lambda$ &- & 0.020 \\
\hline
$\varepsilon$ & 0.001 & 0.001 \\
\hline
\end{tabular}
\end{table}

The performance of the algorithms has been accessed through the use of a normalized mean square error (NMSE) computed for each variable
\begin{equation}
 \mathrm{NMSE}(\hat{\boldsymbol{\theta}})=\frac{ \| \hat{\boldsymbol{\theta}}-\boldsymbol{\theta}\|_F^2}{ \|\boldsymbol{\theta}\|_F^2}
\end{equation}
where $\hat{\boldsymbol{\theta}}$ is the estimated variable and $\boldsymbol{\theta}$ the corresponding ground truth. The NMSE has been measured for the following parameters: the factor proportions $\mathbf{A}_1$ corresponding to the high-uptake region, the remaining factor proportions $\mathbf{A}_{2:K}$, the SBFs affected by the variability $\tilde{\mathbf{M}}_{1}\triangleq [{\mathbf{m}}_{1,1},\ldots,{\mathbf{m}}_{1,N}]$, the non-specific factor TACs $\mathbf{M}_{2:K} \triangleq [\mathbf{m}_2,\ldots,\mathbf{m}_K]$ and finally the variability factor proportion matrix $\mathbf{B}$.


\subsection{Hyperparameter influence}

Considering the significant number of hyperparameters to be tuned in both LMM and SLMM approaches (i.e., $\alpha$, $\beta$, $\lambda$), a full sensitivity analysis is a challenging task, which is further complexified by the non-convex nature of the problem. To alleviate this issue, each parameter has been individually adjusted while the others have been set to zero. Several simulations empirically showed that the result is not very sensitive to the choice of parameters. The parameters have been tuned such that the total percentage of their corresponding term in the overall objective function does not surpass $25$\% of the total value of the function. Given the high level of noise corrupting the PET images, the hyperparameter $\alpha$ associated with the factor proportions has been set so as to reduce the noise impact while avoiding too much smoothing. The factor TAC penalization hyperparameter $\beta$ results from a trade-off between the quality of the initial factor TAC estimates $\mathbf{M}^0$ and the flexibility required by PALM to reach more accurate estimates. Finally the variability penalization $\lambda$ has been tuned to achieve a compromise between the risks of capturing noise into the variability term (i.e., overfitting) and of losing information. While there are more automatized ways to choose the hyperparameter values (e.g., using cross-validation, grid search, random search and Bayesian estimation), these hyperparameter choices have seemed to be sufficient to assess the performance of the proposed method. The hyperparameter values used in LMM and SLMM are finally reported in Table \ref{table:par_noise15}.

\subsection{Results}
\begin{figure}
\begin{center}
\includegraphics[width=\columnwidth,height=0.3\textwidth]{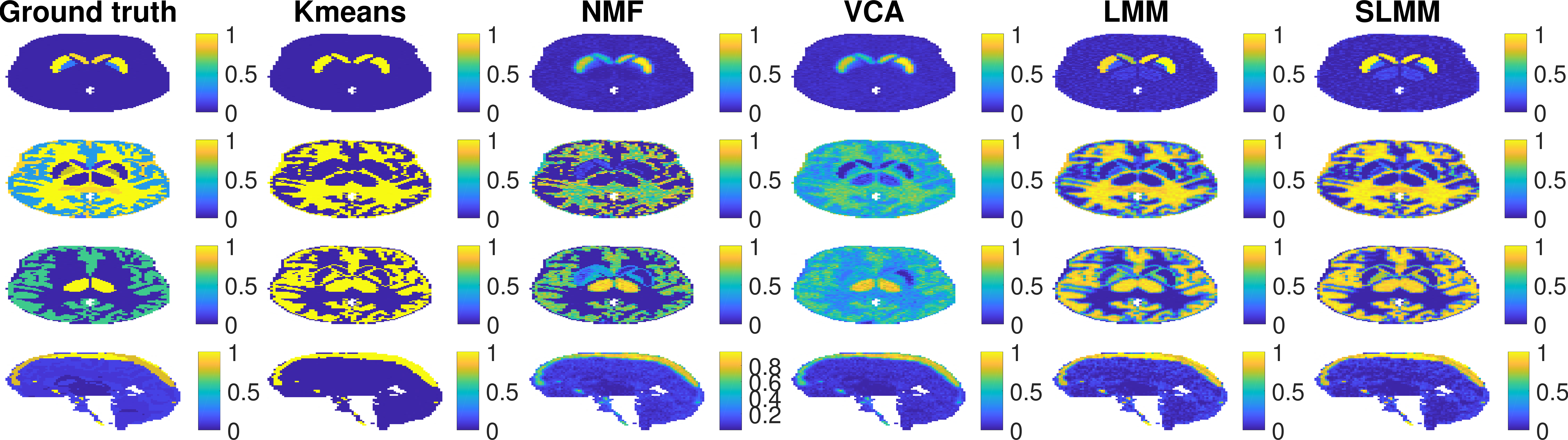}
\caption{Factor proportion maps of the 15th time-frame obtained for SNR=15dB corresponding to the specific gray matter, white matter, gray matter and blood, from top to bottom. The first 3 lines show a transaxial view while the last one shows a sagittal view. All images are in the same scale in $[0,1]$}
\label{fig:A_noise15}
\end{center}
\end{figure}

\begin{figure*}[htpb]
\centering
\includegraphics[width=0.95\textwidth]{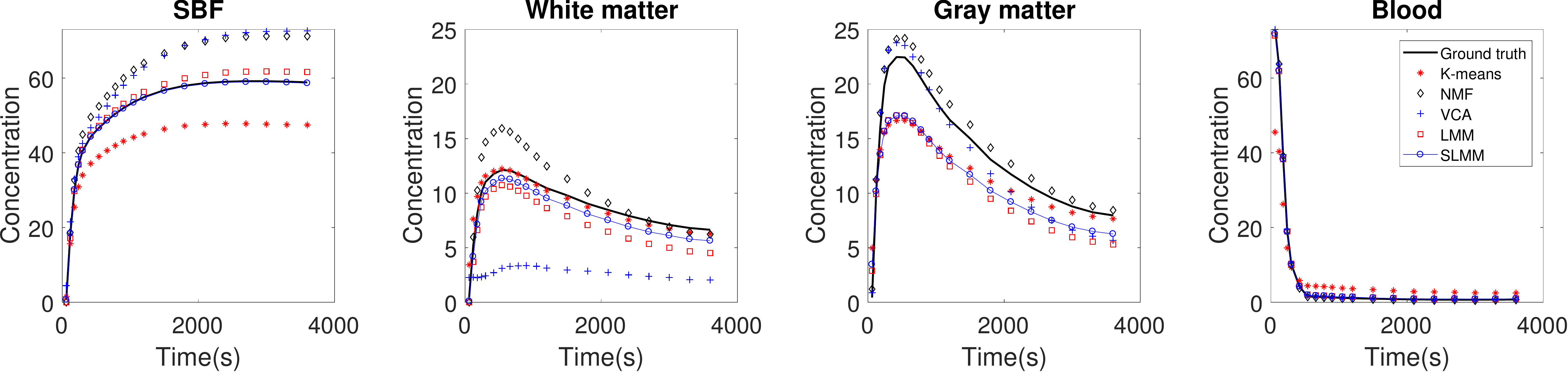}
\caption{TACs obtained for $\mathrm{SNR}=15$dB. For the proposed SLMM algorithm, the represented SBF TAC corresponds to the empirical mean of the estimated spatially varying SBFs ${\mathbf{m}}_{1,1},\ldots,{\mathbf{m}}_{1,N}$.}
\label{fig:M_noise15}
\end{figure*}

\begin{figure}[htbp]
\begin{center}
\includegraphics[width=8cm,height=4cm]{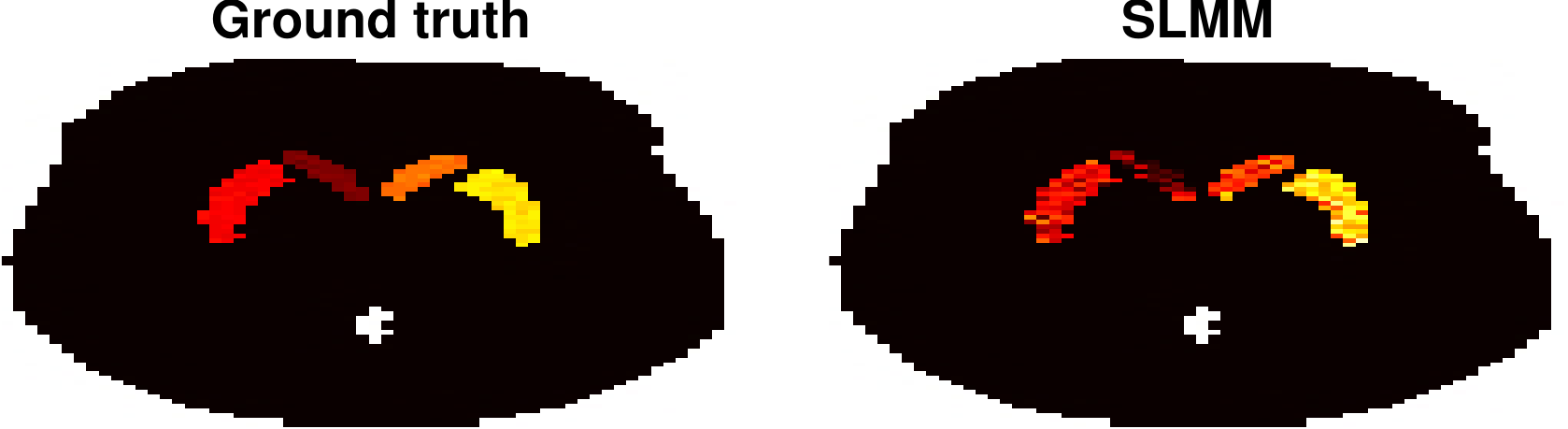}
\caption{Ground-truth (left) and estimated (right) SBF variability.}
\label{fig:B_noise15}
\end{center}
\end{figure}

 The algorithms have been applied to $20$ different realizations of the noise to get reliable performance measures. Table \ref{table:errors_noise15} presents the NMSE averaged over these realizations for all algorithms and variables of interest while Table \ref{table:var_errors_noise15} presents their corresponding variances.
The factor proportion maps recovered by the compared algorithms are shown in Fig. \ref{fig:A_noise15}. Each row corresponds to a specific factor: SBF, white matter factor, non-specific gray matter factor, blood factor (from top to bottom, respectively). The six columns contain the factor proportion maps of the ground truth, and those estimated by K-means, NMF, VCA, LMM and the proposed SLMM (from left to right, respectively).  A visual comparison suggests that the factor proportion maps obtained with LMM and SLMM are more consistent with the expected localization of each factor in the brain than VCA. Meanwhile, they are less noisy than the maps obtained by NMF. The estimated LMM and SLMM proportions maps are closer to the ground truth than both of them, particularly in the region affected by specific binding, as quantitatively shown in Table \ref{table:errors_noise15}. It can also be observed that the factor proportion maps obtained with the proposed SLMM approach present a higher contrast compared to LMM and other approaches, especially in the high-uptake region.

The maps of SLMM are also sharper compared to LMM. Additionally, it is also possible to see that NMF results for white matter are sharper but also more noisy than both LMM and SLMM approaches. However, for the specific gray matter, both LMM and SLMM approaches show sharper estimated factor proportion maps. Note that the sharpness of the factor proportions is not necessarily a good criterion of comparison. Indeed, factor analysis-based methods do expect to recover smooth maps that take into account the spilling part of partial volume effect, which is not considered within deconvolution. The aim of unmixing is not hard-clustering or classification.

The corresponding estimated factor TACs are shown in Fig. \ref{fig:M_noise15} where, for comparison purposes, the SBF depicted for SLMM is the empirical average over the whole set of spatially varying SBFs, as it is also the case for the SBF ground truth TACs. The best estimate of the SBF TAC seems to be obtained by the proposed SLMM approach, for which the TAC has been precisely recovered, as opposed to K-means, VCA and NMF. K-means provide the best estimate of the white matter TAC, closely followed by SLMM while NMF highly overestimates it. The best estimate of the non specific gray matter TAC has been obtained by VCA and NMF, even though it is slightly overestimated. It can be observed that SLMM and LMM have underestimated this factor TAC, which has been compensated with higher values in the corresponding factor proportion map. This underestimation results from the poor initialization of these algorithms by the K-means outputs. A more powerful initialization for the gray matter may provide better results. The factor TAC associated with blood is correctly estimated by SLMM, LMM, VCA and NMF.

The quantitative results in Table \ref{table:errors_noise15} confirm the preliminary findings drawn from the visual inspection of Fig. \ref{fig:A_noise15} and \ref{fig:M_noise15}. The proposed method outperforms all the others for the estimation of $\mathbf{\tilde{M}}_1$, $\mathbf{M}^{2:K}$ and $\mathbf{a}_1$. In particular, SLMM provides a very precise estimation of the mean SBF TAC with an NMSE of $0.9\%$. In Fig. \ref{fig:M_noise15}, the mean of the estimated SBF TACs $\mathbf{m}_{1,1},\ldots,\mathbf{m}_{1,N}$ is very close to the ground truth for LMM and SLMM but the the individual errors computed for each voxel demonstrate better performance obtained by SLMM. It also shows better results than K-means and NMF for $\mathbf{A}_{2:K}$, even though it is less effective but still competitive when compared to LMM and VCA.

\begin{table}
\caption{Mean of Normalized Mean Square Errors of estimated variables for K-means, VCA, NMF, LMM and SLMM with SNR=15dB}
\label{table:errors_noise15}
\centering
\begin{tabular}{|c|c|c|c|c|c|}
\hline
 & $\mathbf{a}_1$ & $\mathbf{A}_{2:K}$ & $\mathbf{\tilde{M}}^1$ & $\mathbf{M}^{2:K}$ & $\mathbf{B}$\\
\hline
K-means             & 0.567          &  0.669         & 0.120 & 0.442 & - \\
\hline
VCA                 & 0.547          & 0.481          & 0.517 & 0.248 & -   \\
\hline
NMF                 & 0.512          & 0.558          & 0.517 & 0.133 & - \\
\hline
LMM                 & 0.437          & \textbf{0.473} & 0.349 & 0.148 & -  \\
\hline
SLMM                & \textbf{0.359} & 0.495          & \textbf{0.009} & \textbf{0.128} & 0.259  \\
\hline
\end{tabular}
\end{table}

\begin{table}
\caption{Variance of Normalized Mean Square Errors of estimated variables for K-means, VCA, NMF, LMM and SLMM with SNR=15dB}
\label{table:var_errors_noise15}
\centering
\begin{tabular}{|c|c|c|c|c|c|}
\hline
 & $\mathbf{a}_1$ & $\mathbf{A}_{2:K}$ & $\mathbf{\tilde{M}}^1$ & $\mathbf{M}^{2:K}$ & $\mathbf{B}$\\
\hline
K-means ($\times 10^{-2}$)        & 0.043          &  0.225         & 0.015 & 6.136 & - \\
\hline
VCA ($\times 10^{-3}$)            & 0.667          & 1.817          & 0.093 & 1.301 & -   \\
\hline
NMF ($\times 10^{-4}$)            & 0.010          & 0.381          & 0.448 & 1.486 & - \\
\hline
LMM ($\times 10^{-6}$)            & 3.791          & 0.043          & 0.602 & 1.469 & -  \\
\hline
SLMM ($\times 10^{-5}$)           & 1.270          & 0.312          & 0.003 & 0.098 & 2.261  \\
\hline
\end{tabular}
\end{table}


Taking into account the SBF variability allows the estimation of $\mathbf{A}_1$ to be improved up to $35\%$. Fig. \ref{fig:B_noise15} compares the actual variability factor proportions and those estimated by the proposed SLMM. This figure shows that the estimated non-zeros coefficients are correctly localized in the $4$ subregions characterized by some SBF variability. These non-zero values seem to be affected by some estimation inaccuracies, mainly due to the deconvolution. However, the estimation error still stays close to $25\%$.

To summarize, in dynamic PET imaging, the TACs associated with the different types of tissues are always highly correlated. Consequently, the conventional LMM approach converges to poor local optima and is only guided by the factor similarity penalization defined by \ref{eq:endmpen2}. On the contrary, SLMM shows better results since it explicitly takes the spatial variability of the SBF into account. VCA associated with SUnSAL shows acceptable results, in particular given the high level of noise affecting the PET images. Finally, as expected, SLMM presents better results for the voxels containing specific gray matter tissue.

\section{Evaluation on Real Data}
\label{sec:real_sim}

\subsection{PET data acquisition}

\begin{figure}[htbp]
\begin{center}
\includegraphics[width=\columnwidth,height=0.3\textwidth]{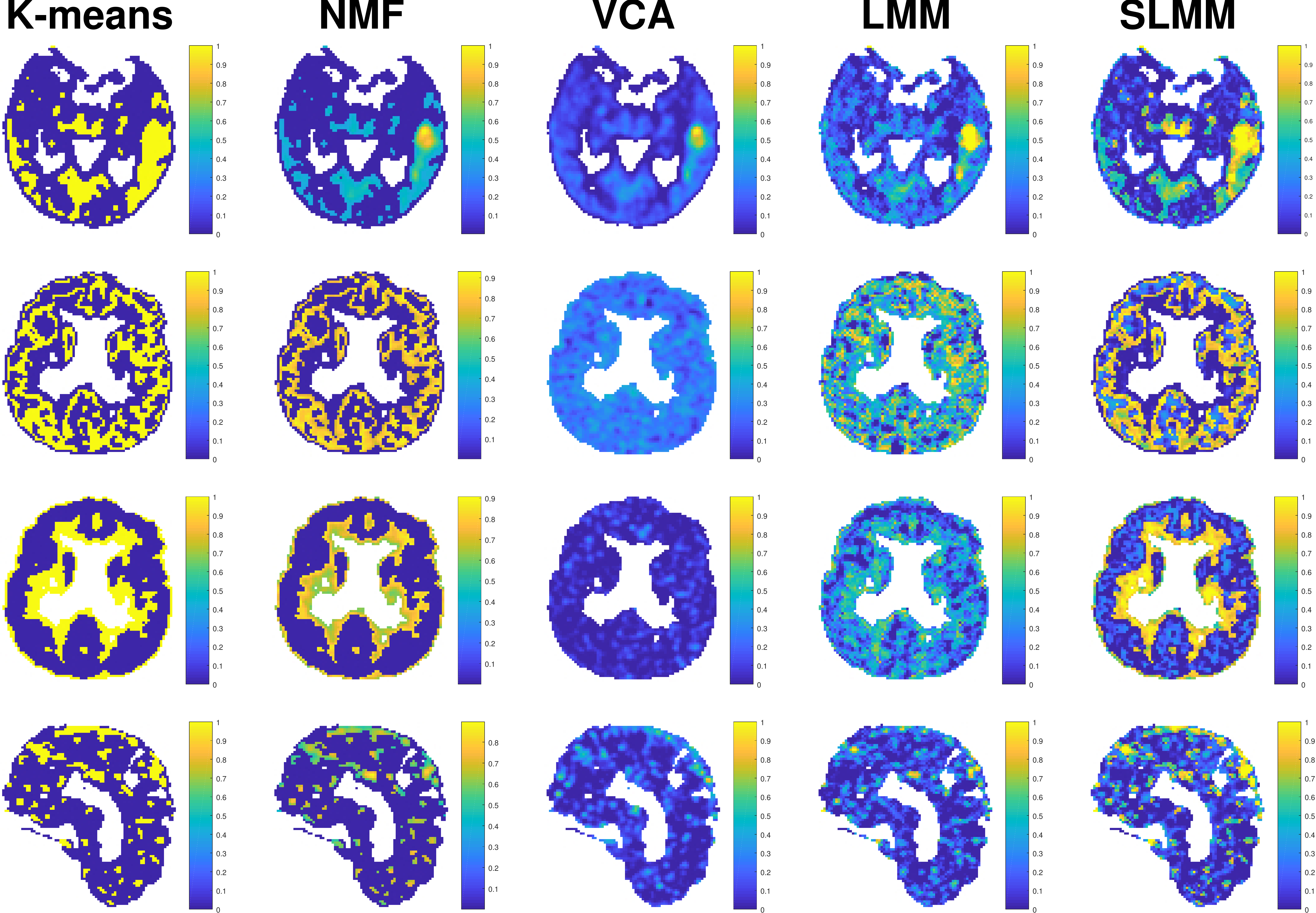}
\caption{factor proportion maps of  the real PET image with [$^{18}F$]DPA-714 of a subject with a stroke. The first 3 lines show a transaxial view while the last one shows a sagittal view. From top to bottom: the specific gray matter, white matter, non-specific gray matter and blood.}
\label{fig:A_real}
\end{center}
\end{figure}

To assess the behavior of the proposed approach when analyzing real dynamic PET images, the different methods have been applied to a dynamic PET image with [18F]DPA-714 of a stroke subject. Cerebral stroke is a severe and frequently occurring condition. While different mechanisms are involved in the stroke pathogenesis, there is an increasing evidence that inflammation, mainly involving the microglial and the immune system cells, account for its pathogenic progression. The [18F]DPA-714 is a ligand of the 18-kDa translocator protein (TSPO) for in vivo imaging, which is a biomarker of neuroinflammation. The subject was examined using an Ingenuity TOF Camera from Philips Medical Systems, seven days after the stroke.

The PET acquisition was reconstructed into a $128 \times 128\times 90$-voxels dynamic PET image with $L=31$ time-frames. The PET scan image registration time ranged from $10$ seconds to $5$ minutes over a $59$ minutes period. The voxel size was of $2 \times 2 \times 2$mm$^3$. As for the experiments conducted on simulated data, the voxel TACs have been assumed to be mixtures of $K=4$ types of elementary TAC: specific binding associated with inflammation, blood, the non-specific gray and white matters. The K-means method was applied to the images to mask the cerebrospinal fluid and to initialize NMF, LMM and SLMM algorithms. A ground truth of the high-uptake tissue was manually labeled by an expert based on a magnetic resonance imaging (MRI) acquisition. The stroke region was segmented on this registered MRI image to define a set of voxels used to learn the variability descriptors $\mathbf{V}$ by PCA. The nominal SBF has been fixed as the empirical average of the corresponding TACs with AUC comprised between the 5th and 10th percentile. The choice to use the average of a percentile instead of the minimum AUC TAC is motivated by the fact that, in this case, the learning set is corrupted by noise and partial volume effects.

\begin{figure*}[htbp]
\includegraphics[width=\textwidth]{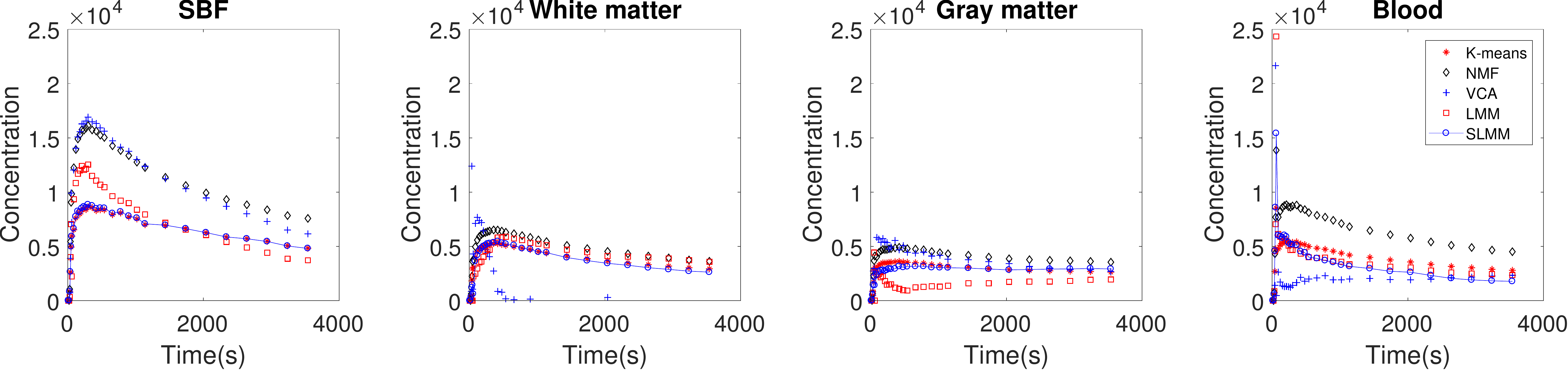}
\caption{TACs obtained by estimation from the real image.}
\label{fig:M_real}
\end{figure*}

\subsection{Results}
Figure \ref{fig:A_real} depicts the factor proportion maps estimated by the compared methods. The corresponding estimated factor TACs are shown in Fig. \ref{fig:M_real}. The LMM and SLMM algorithms estimate four distinct TACs associated with different tissues, as expected. In the first row of Fig. \ref{fig:A_real}, corresponding to SBF, all methods seem to correctly recover the main localization of the stroke area. However, the proposed SLMM approach identifies a significantly larger area. This result seems to be in better agreement with the stroke area identified in the MRI acquisition of the same patient (see $1$st column of Fig. \ref{fig:3D_SBR_real}). Moreover, the specific gray matter factor proportion maps estimated by SLMM and K-means shows high values in the thalamus, which is a region known to present specific binding of [18F]DPA-714. Another remarkable result is the factor proportion maps for the blood. The sagittal view represented in the last row is in the exact center of the brain. Both NMF and SLMM recovers factor proportion maps that are in very good agreement with the superior sagittal sinus vein that passes on the higher part of the brain. On the contrary, VCA estimates two factors which seem to be mixtures of the vein TACs and other region TACs.

Fig. \ref{fig:3D_SBR_real} depicts three different views of the AVC area identified by the expert on MRI acquisition ($1$st column),  the estimated specific gray matter factor proportions ($2$nd-$6$th columns) and the estimated corresponding variability ($7$th column). This figure shows that SLMM provides sharper and more accurate maps, characterized by a larger area with high uptake, as in the MRI ground-truth. This figure also compares the map recovered by K-means, which is used as the initialization of the proposed SLMM, with the final SLMM estimate. It is possible to note an interesting improvement of the final estimate when compared to its initialization. This demonstrates that the method converges to an estimation of the specifically bound gray matter that is more accurate with the proposed model.

\begin{figure*}[htbp]
\begin{center}
\includegraphics[width=0.95\textwidth]{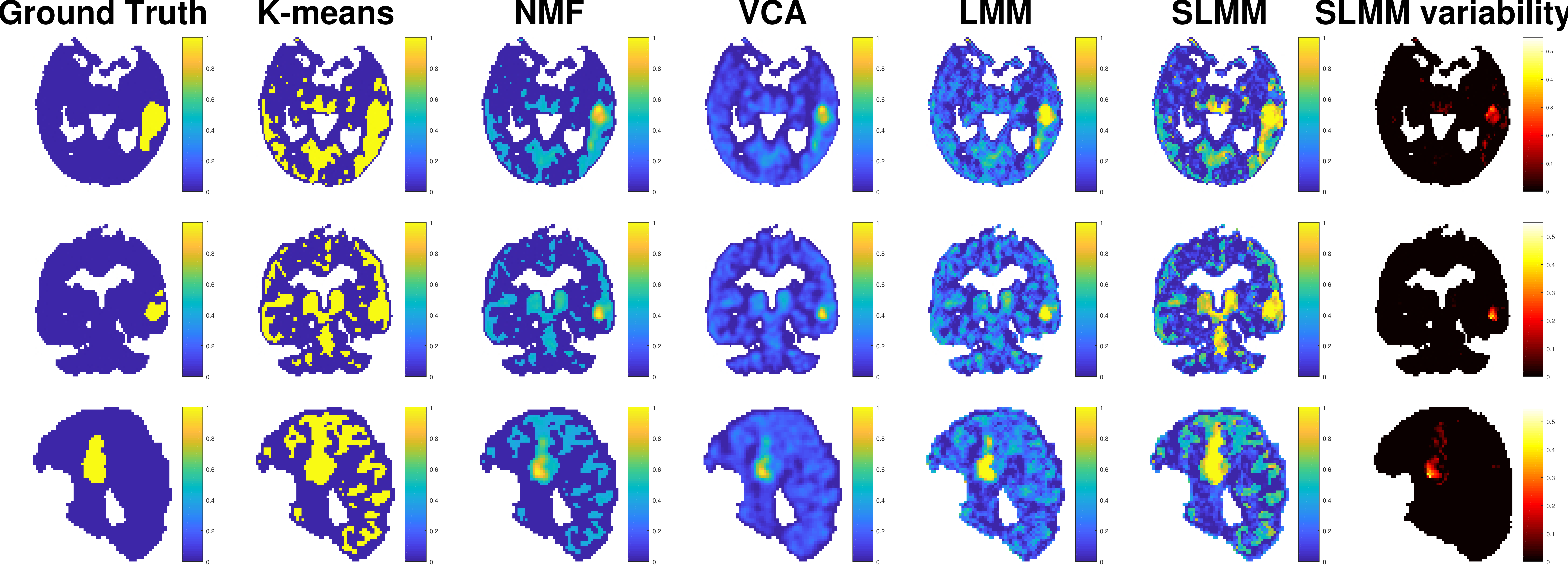}
\caption{From left to right: MRI ground-truth of the AVC area, SBF coefficient maps estimated by K-means, NMF, VCA, LMM, SLMM and SBF variability  estimated by SLMM.}
\label{fig:3D_SBR_real}
\end{center}
\end{figure*}

\section{Conclusion and Future Works}
\label{sec:concl}

This paper introduced a new model to conduct factor analysis of dynamic PET images. It relied on the unmixing concept accounting for specific binding TACs variation. The method was based on the hypothesis that the variations within the SBF can be described by a small number of basis elements and their corresponding proportions per voxel. The resulting optimization  problem is extremely non-convex with highly correlated factors and variability basis elements, which leads to a high number of spurious local optima for the cost function. However, the experiments conducted on synthetic data showed that the proposed method succeeded in estimating this variability, which improved the estimation of the specific binding factor and the corresponding proportions. For the other quantities of interest, the proposed approach compared favorably with state-of-the-art unmixing techniques. The proposed approach has many potential applications in dynamic PET imaging. It could be used for the segmentation of a region-of-interest, classification of the voxels, creation of subject-specific kinetic reference regions or even simultaneous filtering and partial volume correction. Besides exploring such applications of the method, future works should focus on the introduction of a Poisson-fitting measure of divergence used in the cost function, e.g. Kullback-Leibler, to better model noise frequently encountered in low rate PET data.

    \bibliographystyle{ieeetran_shorten} 
\bibliography{strings_all_ref,bibli}

\end{document}